\documentclass{article}

\usepackage[preprint]{neurips_2024}

\usepackage[utf8]{inputenc} 
\usepackage[T1]{fontenc}    
\usepackage{hyperref}       
\usepackage{url}            
\usepackage{booktabs}       
\usepackage{amsfonts}       
\usepackage{nicefrac}       
\usepackage{microtype}      
\usepackage{xcolor}         

\usepackage{amsmath}             
\usepackage{amssymb}             
\usepackage{amsthm}              
\usepackage{graphicx}            
\usepackage{subcaption}          
\usepackage{float}               
\usepackage{multirow}            
\usepackage{tabularx}            
\usepackage{siunitx}             

\title{SMOGAN: Synthetic Minority Oversampling with \\ 
GAN Refinement for Imbalanced Regression}

\author{%
  Shayan Alahyari \\
  Department of Computer Science\\
  Western University\\
  London, Ontario, Canada \\
  \texttt{salahya@uwo.ca} \\
  \And
  Mike Domaratzki \\
  Department of Computer Science\\
  Western University\\
  London, Ontario, Canada \\
  \texttt{mdomarat@uwo.ca} \\
}

\begin{document}

\maketitle

\begin{abstract}
Imbalanced regression refers to prediction tasks where the target variable is skewed. This skewness hinders machine learning models, especially neural networks, which concentrate on dense regions and therefore perform poorly on underrepresented (minority) samples. Despite the importance of this problem, only a few methods have been proposed for imbalanced regression. Many of the available solutions for imbalanced regression adapt techniques from the class imbalance domain, such as linear interpolation and the addition of Gaussian noise, to create synthetic data in sparse regions. However, in many cases, the underlying distribution of the data is complex and non-linear. Consequently, these approaches generate synthetic samples that do not accurately represent the true feature-target relationship. To overcome these limitations, we propose SMOGAN, a two-step oversampling framework for imbalanced regression. In Stage 1, an existing oversampler generates initial synthetic samples in sparse target regions. In Stage 2, we introduce DistGAN, a distribution-aware GAN that serves as SMOGAN's filtering layer and refines these samples via adversarial loss augmented with a Maximum Mean Discrepancy objective, aligning them with the true joint feature-target distribution. Extensive experiments on 23 imbalanced datasets show that SMOGAN consistently outperforms the default oversampling method without the DistGAN filtering layer.
\end{abstract}

\textbf{Keywords:} GAN, Imbalanced, Regression, Generative AI, Unsupervised learning

\section{Introduction}
\begin{figure}[t]
  \centering
  \includegraphics[width=0.6\columnwidth]{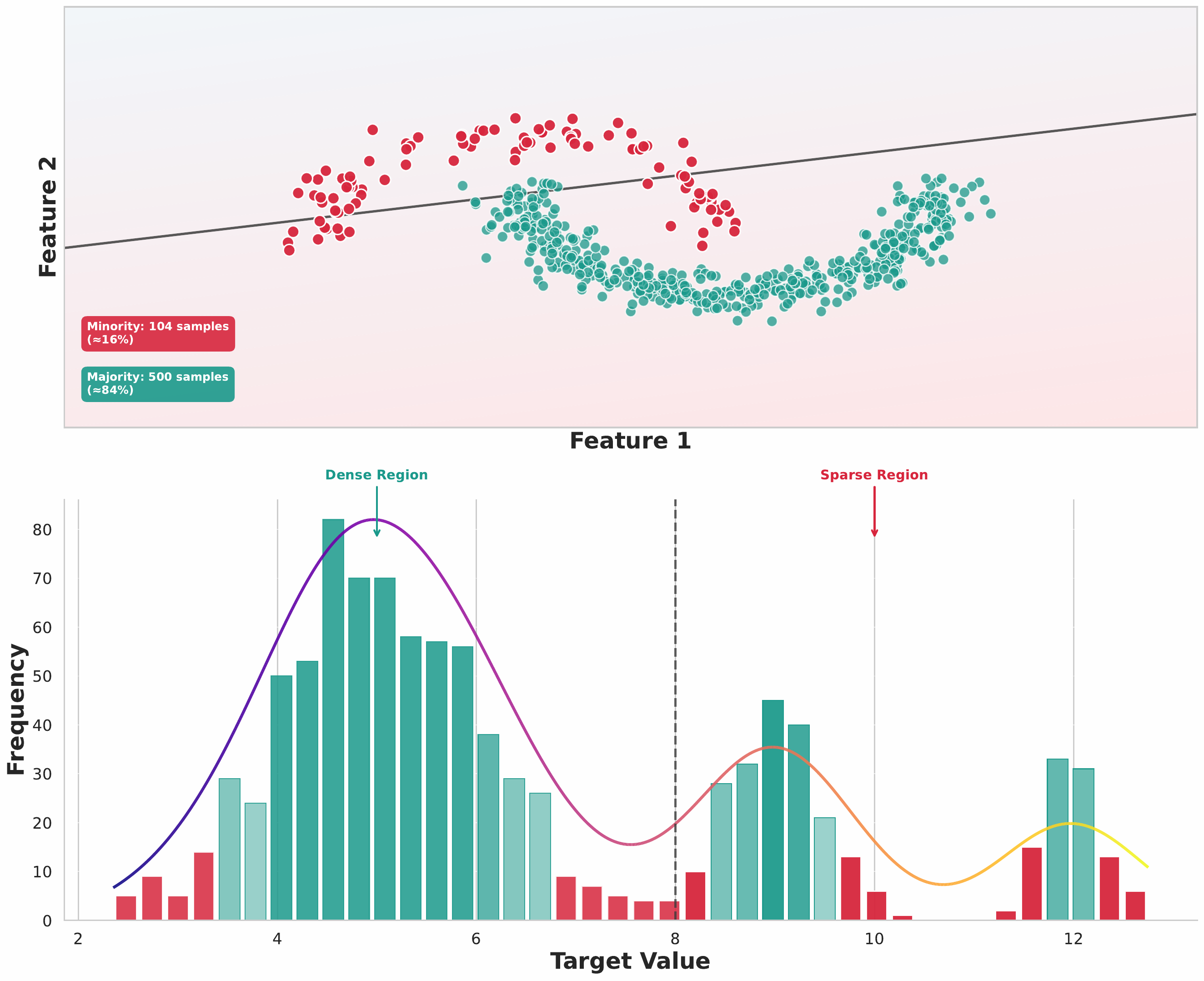}
  \caption{The top panel shows a classification problem with easily identifiable minority (red) and majority (green) classes. The bottom panel shows an imbalanced regression problem where target values in the sparse region (red regions) are underrepresented, making them more difficult to detect and accurately predict.}
  \label{fig:imbalance_problem}
\end{figure}

Imbalanced data in regression tasks that predict continuous outcomes can compromise model performance \citep{krawczyk2016}. In such settings, some target ranges, particularly minority samples, have very few observations \citep{branco2016,torgo2013}. Standard regression models tend to focus on densely populated regions of the target space, resulting in poor predictions for underrepresented values \citep{chawla2004,branco2016}. Yet, accurately forecasting these uncommon targets is essential for many practical applications \citep{torgo2013}.

Conversely, in both binary and multi-class classification, imbalanced data occurs when some classes have far fewer samples than others. This leads classifiers to focus on the majority classes while missing the sparse patterns of minority classes. Effective techniques for addressing class imbalance are essential to ensure that all classes, even those with fewer examples, are accurately recognized \citep{he2009,chawla2002,buda2018,liu2009}. Critical applications like fraud detection, medical diagnostics, and fault detection are particularly affected, as important minority classes might be missed \citep{haixiang2017,johnson2019}.

Imbalanced regression impacts many critical real-world applications where accurately predicting rare but high-impact events in the tails of skewed distributions is paramount. For instance, In air quality forecasting, infrequent high ozone concentration events are systematically underestimated by SVR models trained on abundant moderate values, posing health risk underpredictions \citep{zhen2025weighted}. In intensive care, predicting prolonged ventilation days from skewed ICU datasets is hampered by scarce samples of long-duration cases, leading to suboptimal resource planning \citep{radovanovic2025tackling}. In precision agriculture, vineyard yield datasets feature extreme low- and high-yield blocks that are poorly represented, prompting conditional UNet–ConvLSTM architectures with zonal weighting to address these extremes \citep{kamangir2024large}. In chemical synthesis planning, high-yield reactions are rarer yet more valuable, yet existing prediction models overlook this imbalance and underperform on critical high-yield bins \citep{ma2024revisiting}. These cases underscore the inadequacy of conventional regression and the need for specialized imbalanced-regression methodologies.

Figure~\ref{fig:imbalance_problem} shows the difference between imbalanced classification and regression. In classification, minority classes are easy to spot because the labels are distinct and countable. In regression, however, few-sample areas occur within continuous outcomes, usually at the extremes. Unlike discrete labels, these underrepresented regions depend on the distribution's shape. Consequently, standard methods from classification do not work, making it challenging to detect and address these sparse areas.

\begin{figure}[t]
  \centering
  \includegraphics[width=0.5\columnwidth]{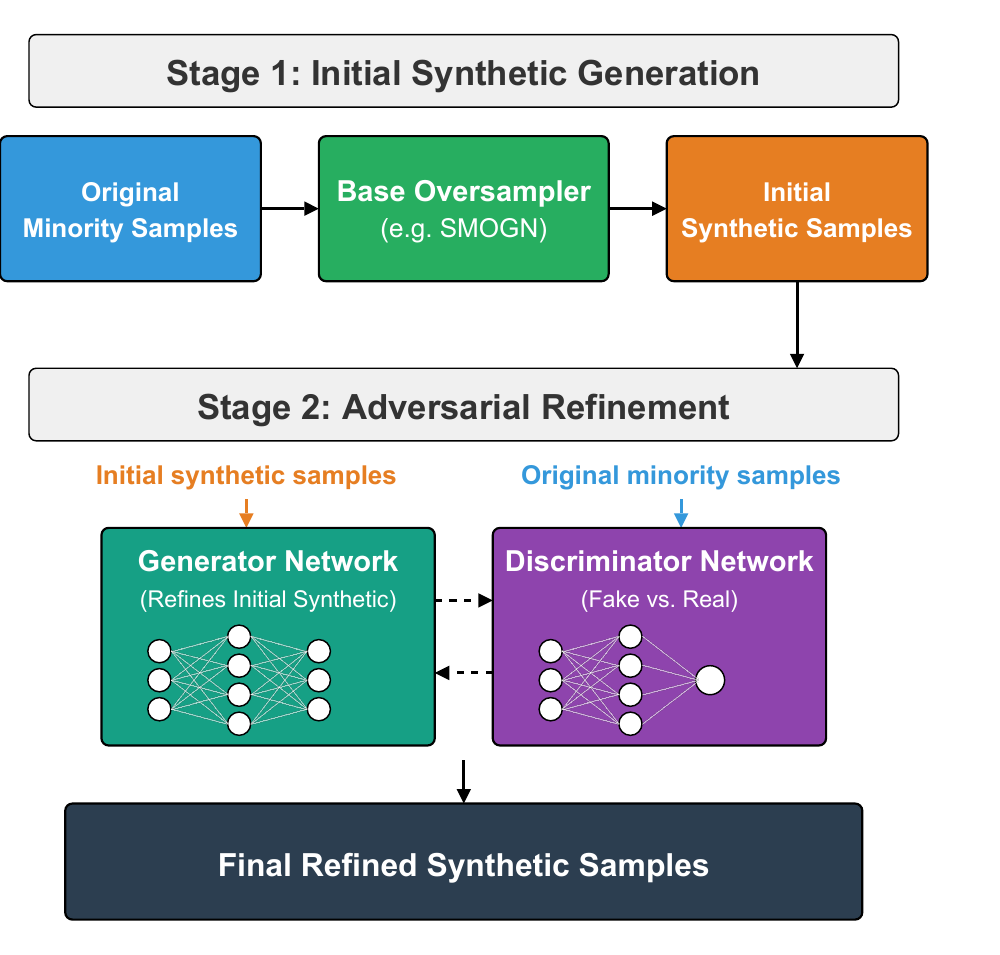}
  \caption{SMOGAN workflow: initial oversampling of minority samples is followed by adversarial refinement to produce realistic synthetic examples.}
  \label{fig:SMOGAN_process}
\end{figure}

Existing oversampling methods for imbalanced regression often oversimplify the problem and overlook the complex underlying structure of continuous target distributions. To address this limitation, we propose SMOGAN, a two-step framework that integrates DistGAN (a distribution-aware generative adversarial network \citep{goodfellow2014generative}) as an inbuilt filtering layer to refine synthetic samples and preserve real data fidelity. As illustrated in Figure~\ref{fig:SMOGAN_process}, in Stage 1 (Initial Synthetic Generation), a base oversampler (e.g., SMOGN) produces initial synthetic samples in sparse target regions. In Stage 2 (Adversarial Refinement), these candidates are passed through DistGAN’s generator and discriminator: the generator refines them by optimizing a combined adversarial loss augmented with Maximum Mean Discrepancy (MMD) to minimize kernel-based distribution discrepancies, while the discriminator, trained exclusively on real minority samples, filters out-of-distribution outputs. We conduct comprehensive experiments on 23 benchmark imbalanced regression datasets. We extend our evaluation across diverse application domains to illustrate the effectiveness and generalizability of the proposed SMOGAN framework. Importantly, SMOGAN’s modular design allows the initial oversampler in Stage 1 to be replaced with any data-level technique of the user’s choice, enabling seamless integration of new or domain-specific methods with DistGAN’s refinement process.

\section{Related Work}

\subsection{Class imbalance}

This section reviews key methods for addressing class imbalance in classification, as these techniques serve as the basis for several regression approaches we evaluate. The foundational work by \citet{chawla2002} presented SMOTE (Synthetic Minority Over-sampling Technique), the first data-level solution to class imbalance. SMOTE creates artificial minority class examples \(\tilde{\mathbf{x}}\) through linear interpolation between a minority sample \(\mathbf{x}_i\) and one of its \(k\) nearest minority neighbors \(\mathbf{x}_{i,\mathrm{nn}}\):
\[
\tilde{\mathbf{x}}
= \mathbf{x}_i
+ \lambda \bigl(\mathbf{x}_{i,\mathrm{nn}} - \mathbf{x}_i\bigr),
\quad
\lambda \sim \mathcal{U}(0,1).
\]

This method established itself as the standard approach for managing imbalanced datasets, inspiring numerous extensions and modifications. The core interpolation mechanism of SMOTE provides crucial insights for understanding its regression counterparts, particularly SMOTER and SMOGN, which we analyze in our experiments \citep{torgo2013, branco2017}.


Beyond SMOTE and its derivatives, researchers have applied generative adversarial networks (GANs) to tackle class imbalance problems. \citet{mariani2018bagan} introduced BAGAN, which combines an autoencoder-based initialization with class-conditional generation to create varied and realistic minority class images. In the tabular domain, \citet{tanaka2019dataaugmentation} showed that synthetic data from GANs could effectively substitute real training samples while enhancing minority class detection rates. For mixed-type tabular data, \citet{engelmann2020conditional} proposed a conditional WGAN architecture incorporating gradient penalties, auxiliary classification objectives, Gumbel-softmax activation for discrete variables, and cross-layer connections.

Recent work has also integrated traditional oversampling with deep generative models. \citet{sharma2022smotified} combined SMOTE's interpolation strategy with GAN-based generation, using interpolated samples to guide the generator toward more diverse synthetic minority instances to tackle class imbalance. Applications extend to sequential data, where \citet{jiang2019gan} developed an encoder-decoder-encoder GAN framework for time-series anomaly detection using joint reconstruction and adversarial objectives. Similarly, \citet{lee2019ganids} addressed rare attack detection in network intrusion systems by generating synthetic attack patterns with GANs and subsequently retraining random forest classifiers.

\subsection{Imbalanced regression}

Imbalanced regression has received far less attention than classification imbalance. \citet{torgo2007utility} and \citet{torgo2013} developed SMOTER, the first data-level method for regression. SMOTER uses a relevance function $\phi(y)$ to find rare target values, then undersamples common cases and interpolates between rare ones.

\citet{branco2017} improved SMOTER by adding Gaussian noise (SMOGN) to create more varied synthetic samples. They later developed WERCS \citep{branco2019}, which uses $\phi(y)$ to decide whether to oversample or undersample each instance using random selection or noise. \citet{camacho2022} developed G-SMOTE, which considers the geometry of target values when interpolating. \citet{camacho2024} replaced SMOTER's fixed thresholds with continuous weights in WSMOTER. \citet{alahyari2025ldao} proposed LDAO, which clusters the joint feature-target space and applies kernel density estimation within each cluster to generate synthetic samples that preserve local distribution characteristics. \citet{alahyari2025regression} used Mahalanobis-GMM detection with GAN-based generation for imbalanced regression, eliminating manual threshold selection while preserving statistical properties of minority samples. \citet{moniz2018} combined SMOTE with boosting in SMOTEBoost, showing that oversampling works well with ensemble methods for regression.

Researchers have also developed algorithmic methods for imbalanced regression. Cost-sensitive learning weights errors on rare targets more heavily than common ones \citep{zhou2010,elkan2001,domingos1999}. \citet{steininger2021} created DenseLoss, which gives more importance to low-density targets in the loss function. \citet{yang2021} used label-distribution smoothing in neural networks to handle imbalance. \citet{ren2022} introduced Balanced MSE, which scales errors based on how rare each target value is.

New evaluation metrics have been developed specifically for imbalanced regression. \citet{ribeiro2020} created SERA (Squared Error Relevance Area), which weights errors based on target rarity. Unlike RMSE, which treats all errors equally, SERA gives more weight to errors on rare values by integrating across different relevance thresholds. Standard metrics hide poor performance on rare samples because errors on common samples dominate the overall score. This makes metrics like SERA essential for evaluating imbalanced regression methods properly.

\section{SMOGAN method}

\begin{figure*}[!t]
  \centering
  \includegraphics[width=\columnwidth]{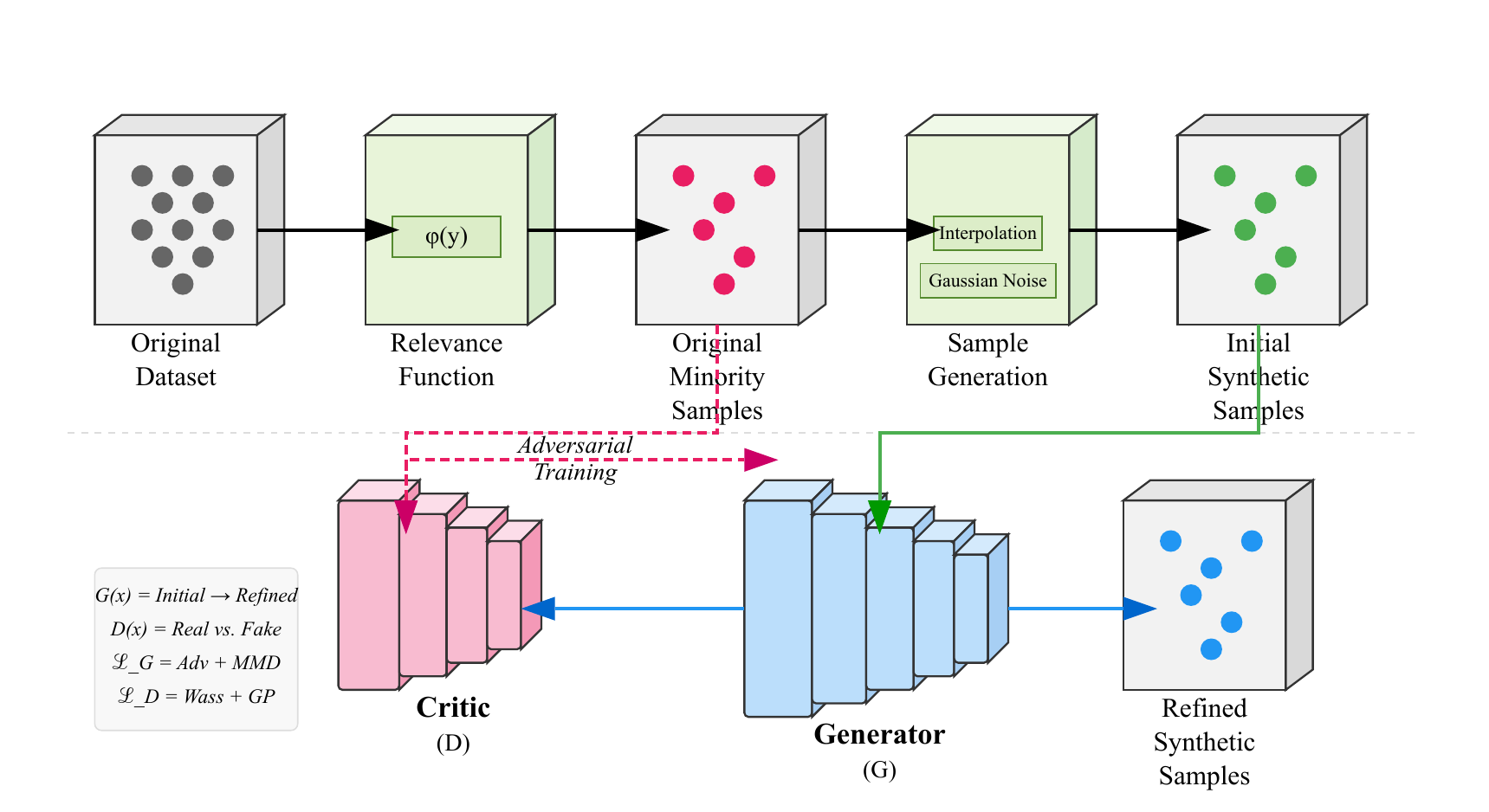}
  \caption{SMOGAN two-stage framework: Stage 1 applies SMOGN-based oversampling to generate initial minority samples; Stage 2 uses a WGAN-GP with MMD regularization to adversarially refine those samples.}
  \label{fig:SMOGAN_process2}
  
\end{figure*}

\subsection{Stage 1: Initial Synthetic Generation}
\label{ssec:smogn_stage1}
We base our initial oversampling on SMOGN (SmoteR and Gaussian noise) \citep{branco2017}, a popular regression oversampler, chosen for its ability to both interpolate and inject variability in sparse target regions. We initialize by generating an initial pool of synthetic samples using SMOGN. Each target $y$ in the original data set is assigned a relevance score $\phi(y)\in[0,1]$, labeling values within the interquartile range as unimportant and those in the tails as increasingly important via a simple sigmoid function.  We then split the data into rare ($\phi(y)\ge t_R$) and normal ($\phi(y)<t_R$) cases, using a relevance threshold $t_R$ (typically 0.8).  

Focusing only on the rare set, each seed sample $(\mathbf x_i,y_i)$ spawns synthetic neighbours as follows:
1. Select one of its $k$ nearest rare neighbours $(\mathbf x_j,y_j)$ and compute the Euclidean distance $d_{ij}=\|\mathbf x_j-\mathbf x_i\|$.
2. Let $d^\star$ be half the median of $k$ distance.  If $d_{ij}<d^\star$, interpolate (SmoteR):
   \[
   \tilde{\mathbf x} = \mathbf x_i + u(\mathbf x_j-\mathbf x_i),\quad
   \tilde y = y_i + u(y_j-y_i),\quad
   u\sim\mathcal U(0,1).
   \]
   Otherwise, apply Gaussian jitter:
   \[
   \tilde{\mathbf x} = \mathbf x_i + \varepsilon,\quad
   \tilde y = y_i + \eta,\quad
   \varepsilon\sim\mathcal N(0,\sigma^2I),\;
   \eta\sim\mathcal N(0,\sigma^2),
   \]
   with $\sigma=\min(0.02, d^\star)$.  

Repeating this process $N$ times per seed produces the initial synthetic pool:
\[
\mathcal D_{\mathrm{syn}}^{(0)} = \{(\tilde{\mathbf x},\tilde y)\},
\]
which populates sparse target regions and serves as input to Stage 2 for adversarial refinement.

\subsection{Stage 2: Adversarial Refinement with DistGAN}

Having generated the initial synthetic pool $\mathcal D_{\mathrm{syn}}^{(0)}$, we refine it by training a Wasserstein GAN with gradient penalty (WGAN-GP) \citep{arjovsky2017wasserstein,gulrajani2017improved}. We choose WGAN-GP because it provides stable training dynamics and generates diverse, high-quality samples without mode collapse, critical for preserving the complex structure of minority regions in tabular data. In the GP variant, weight clipping is replaced by a gradient-penalty term that enforces the $1$-Lipschitz constraint, yielding smoother gradients and more stable training. On top of this, the generator loss incorporates an RBF-kernel MMD term to align all moments between refined and real minority samples. Figure~\ref{fig:SMOGAN_process2} illustrates this two-stage adversarial refinement workflow.

\subsubsection{Critic objective}

The critic \(D\) must assign higher scores to real samples than to generated ones while remaining 1-Lipschitz.  We enforce the Lipschitz constraint via a gradient penalty on random interpolates
\[
\hat u = \epsilon\,z + (1-\epsilon)\,G(x),
\quad
\epsilon\sim\mathcal{U}(0,1),
\]
with \(z\sim\mathcal D_{\mathrm{real}}\) and \(x\sim\mathcal D_{\mathrm{syn}}^{(0)}\).  Here, \(G\) is the generator network, and \(G(x)\) denotes the refined synthetic sample produced by feeding the initial SMOGN-generated point \(x\) through the generator.

When updating the critic, we minimize the following loss with respect to its parameters~\(\theta_D\):
\begin{align*}
\mathcal L_D
&=
\underbrace{\mathbb E_{x\sim\mathcal D_{\mathrm{syn}}^{(0)}}\!\bigl[D(G(x))\bigr]
- \mathbb E_{z\sim\mathcal D_{\mathrm{real}}}\!\bigl[D(z)\bigr]}_{\text{Wasserstein objective}}
+\,\lambda_{\mathrm{gp}}\;
\underbrace{\mathbb E_{\substack{z,x,\epsilon}}
  \bigl(\|\nabla_{\hat u}D(\hat u)\|_2 - 1\bigr)^2}_{\text{gradient penalty}}.
\end{align*}
We set \(\lambda_{\mathrm{gp}} = 10\), following the hyper-parameter recommendation in the original WGAN-GP paper ~\citep{gulrajani2017improved}.

\subsubsection{Generator objective}
\label{sssec:gen}

The generator \(G\) refines each fixed seed \(x\in\mathcal D_{\mathrm{syn}}^{(0)}\) and is trained to
(i) lower the critic score of its outputs and (ii) minimize a kernel Maximum Mean Discrepancy
against real minority samples:
\begin{align*}
\mathcal L_G
&=
\underbrace{-\mathbb E_{x\sim\mathcal D_{\mathrm{syn}}^{(0)}}
            D\bigl(G(x)\bigr)}_{\text{adversarial term}}
+\,\alpha\;
\underbrace{\widehat{\mathrm{MMD}}^{2}
           \bigl(\{G(x_i)\},\{z_j\}\bigr)}_{\text{moment-matching term}},
\end{align*}
where \(\{z_j\}_{j=1}^{M}\subset\mathcal D_{\mathrm{real}}\).
The unbiased batch estimator is
{\scriptsize
\begin{align*}
\widehat{\mathrm{MMD}}^{2}(X,Y)
&=
\underbrace{\frac{1}{N(N-1)}\sum_{i\neq i'}k(x_i,x_{i'})}_{\substack{\text{pairwise terms within }X\\(|X|=N)}}
+\underbrace{\frac{1}{M(M-1)}\sum_{j\neq j'}k(z_j,z_{j'})}_{\substack{\text{pairwise terms within }Y\\(|Y|=M)}}\\
&\quad
-2\underbrace{\frac{1}{NM}\sum_{i=1}^N\sum_{j=1}^M k(x_i,z_j)}_{\text{cross‐sample terms between }X\text{ and }Y}\,.
\end{align*}
}
with the Gaussian kernel
\[
k(u,v)=\exp\!\bigl(-\|u-v\|^{2}/(2\sigma^{2})\bigr),\qquad
\sigma=\operatorname{median}\bigl\{\|u-v\|\bigr\}.
\]
We set \(\alpha=1\); in practice this hyper-parameter can be tuned if either term
dominates.

\subsubsection{Output of stage 2}
After convergence, the refined pool is the generator’s output:
\[
\mathcal D_{\mathrm{syn}}^{(\mathrm{ref})}
=\bigl\{\,G(x)\;:\;x\in\mathcal D_{\mathrm{syn}}^{(0)}\bigr\}.
\]
These samples replace $\mathcal D_{\mathrm{syn}}^{(0)}$ in all downstream
training and evaluation.

\subsubsection{Network architecture}
\label{sssec:arch}
Both the generator \(G\) and critic \(D\) are 4-layer MLPs with ReLU activations (widths: \(d\to128\to256\to128\to d\) for \(G\); \(d\to128\to256\to128\to1\) for \(D\)).

\section{Evaluation methodology}

We benchmark SMOGAN against SMOGN, the current state-of-the-art imbalanced regression oversampler that extends SMOTER with Gaussian noise, using a diverse collection of benchmark datasets. SMOGN requires specifying a relevance function \(\phi(y)\) and a relevance threshold \(t_R\); in our experiments, we employ the default threshold \(t_R=0.8\), as recommended by the authors across all datasets. First, we evaluate SMOGN with its default parameter settings. Then we integrate the DistGAN layer into SMOGN to form SMOGAN, with the primary goal of refining the synthetic sample pool and demonstrating improved sample quality under this configuration. We further evaluate SMOGAN against a GAN‑only approach that skips the initial SMOGN oversampling step. We also compare it to a baseline with no oversampling and to other state‑of‑the‑art methods, including random oversampling (RO), G‑SMOTE, and WSMOTER.

\subsection{Datasets}

We evaluated our method using 23 datasets from three sources: the Keel repository \citep{alcala2011}, the DataSets‑IR collection \citep{branco2019}, and the imbalancedRegression repository \citep{avelino2024}. These datasets span multiple domains and vary in size, dimensionality, and degree of imbalance, making them benchmarks for the rigorous evaluation of imbalanced regression methods and enabling fair comparisons. Table~\ref{tab:dataset-characteristics} shows the number of instances and features for each dataset.

\begin{table}[t]
  \centering
  \scriptsize
  \setlength{\tabcolsep}{3pt}
  \caption{Dataset characteristics showing number of instances and features.}
  \label{tab:dataset-characteristics}
  \resizebox{\columnwidth}{!}{%
  \begin{tabular}{lrr | lrr}
    \toprule
    dataset           & instances & features & dataset           & instances & features \\
    \midrule
    abalone           & 4,177     & 8        & heat              & 7,400     & 11        \\
    acceleration      & 1,732     & 14       & house             & 22,784    & 16        \\
    airfoild          & 1,503     & 5        & kdd               & 316        & 18        \\
    analcat           & 450        & 11       & lungcancer        & 442        & 24        \\
    available\_power  & 1,802     & 15       & maximal\_torque   & 1,802     & 32        \\
    boston            & 506        & 13       & meta              & 528        & 65        \\
    california        & 20,640    & 8        & mortgage          & 1,049     & 15        \\
    cocomo            & 60         & 56       & sensory           & 576        & 11        \\
    compactiv         & 8,192     & 21       & treasury          & 1,049     & 15        \\
    cpu               & 8,192     & 12       & triazines         & 186        & 60        \\
    debutanizer       & 2,394     & 7        & wine\_quality     & 1,143     & 12        \\
    fuel              & 1,764     & 37       &                   &            &           \\
    \bottomrule
  \end{tabular}%
  }
\end{table}

\definecolor{headercolor}{RGB}{70, 130, 180}
\definecolor{rowcolor}{RGB}{240, 248, 255}

\subsection{Metrics}
We evaluate our method using multiple metrics that measure performance on both frequent and rare target values.

\subsubsection{Root mean square error (RMSE)}
Root Mean Square Error (RMSE) measures the overall prediction accuracy by calculating 
the square root of the average squared difference between predicted values \(\hat{y}_i\) and actual observations \(y_i\):
\begin{equation*}
   \text{RMSE} 
   = \sqrt{\frac{1}{n}\sum_{i=1}^{n}(y_i - \hat{y}_i)^2},
   \quad
\end{equation*}

\subsubsection{Squared error-relevance area (SERA)}
SERA is a metric that provides a flexible way to evaluate models under non-uniform domain preferences. 
Let \(\phi(y)\in[0,1]\) be a relevance function that assigns higher scores to rare target values \citep{ribeiro2020}. Then for any relevance 
threshold \(t\), let
\[
D^t 
= \{(x_i,y_i)\mid \phi(y_i)\ge t\},
\]
and define
\begin{equation*}
  SER_t 
  = \sum_{(x_i,y_i)\in D^t} \bigl(\hat{y}_i - y_i\bigr)^{2}.
\end{equation*}

SERA then integrates this quantity over all \(t\in[0,1]\):
\begin{equation*}
  SERA
  = \int_{0}^{1} SER_t \,dt
  = \int_{0}^{1} \sum_{(x_i,y_i)\in D^t} \bigl(\hat{y}_i - y_i\bigr)^{2}\,dt.
\end{equation*}

\subsubsection{Precision, recall and f‑measure for regression}

Precision and recall were first adapted to regression to assess how well predictions focus on user‑specified rare regions rather than minimizing overall error \citep{torgo2009, ribeiro2011a, branco2019}.

Let \(t_R\) be the threshold defining “rare” values, and \(U(\hat y_i,y_i)\in[-1,1]\) a utility rewarding accurate placement within that region. Then:

\begin{align*}
  \mathrm{prec}_\phi &= \frac{\sum_{\phi(\hat y_i)>t_R}\bigl(1 + U(\hat y_i,y_i)\bigr)}
                          {\sum_{\phi(\hat y_i)>t_R}\bigl(1 + \phi(\hat y_i)\bigr)},\\
  \mathrm{rec}_\phi  &= \frac{\sum_{\phi(y_i)>t_R}\bigl(1 + U(\hat y_i,y_i)\bigr)}
                          {\sum_{\phi(y_i)>t_R}\bigl(1 + \phi(y_i)\bigr)},\\
  F_{\phi_1}        &= \frac{2\,\mathrm{prec}_\phi\,\mathrm{rec}_\phi}
                          {\mathrm{prec}_\phi + \mathrm{rec}_\phi}.
\end{align*}

\subsection{Experimental Framework}
For each dataset, we employed 25 different random splits. Each split partitioned the dataset into 80\% for training and 20\% for testing. Models were trained on the 80\% portion and evaluated on the remaining 20\%. For our evaluation metrics, we utilized the SERA implementation from the ImbalancedLearningRegression Python package \citep{wu2022}. The SMOGN method was implemented using the package developed by \citep{kunz2020}.  

We trained a TabNet regressor \citep{arik2021tabnet} on each training set, holding out 20\% for validation. Training ran for up to 1000 epochs with the Adam optimizer (learning rate = 0.0001). The model checkpoint with the lowest validation RMSE was retained and evaluated on the held-out test set. TabNet has six transformer blocks (one initial block plus five decision blocks). Each block contains four fully connected layers of 64 neurons, five attention layers of 16 neurons each, and finally, a single output layer with 1 neuron. We also trained a Random Forest regressor using scikit-learn’s default implementation on the full training set, and then evaluated it directly on the held-out test split \citep{breiman2001random}.

\section{Qualitative Evaluation}
In this section, using the Boston dataset as a representative example, we present the results of a qualitative evaluation comparing synthetic minority samples generated by SMOGN with threshold \(t_R=0.8\) and those refined by DistGAN filtration (SMOGAN) to real minority samples, demonstrating which approach best captures the true underlying distribution of the data.

\begin{figure}[!t]
  \centering
  \begin{minipage}[t]{0.44\linewidth}
    \centering
    \includegraphics[width=\linewidth]{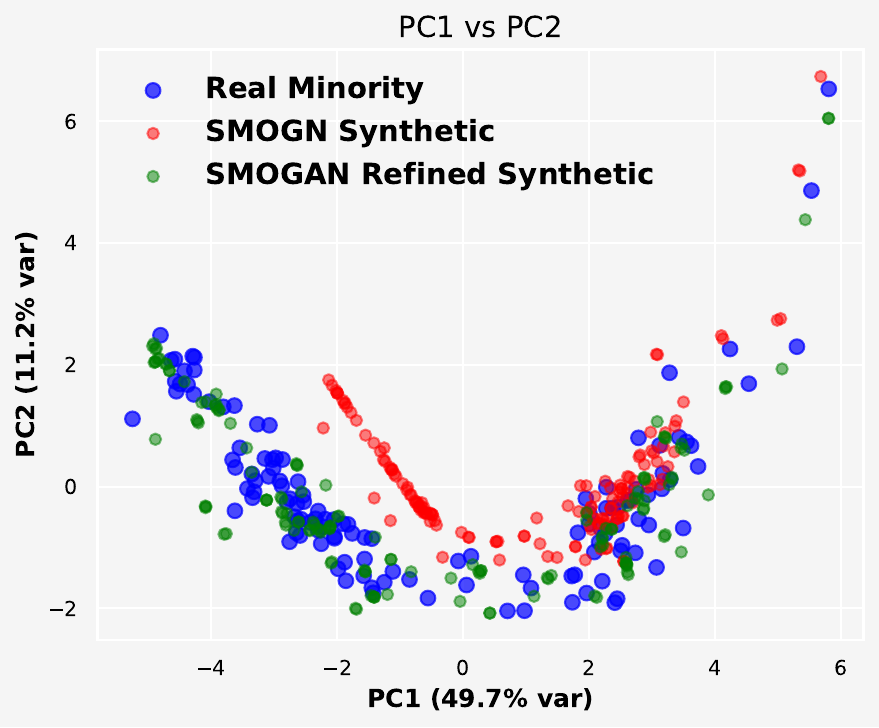}
    \par\smallskip
    {\scriptsize\textbf{(a)} PC1 vs PC2}
    \label{fig:pc1_pc2}
  \end{minipage}
  \begin{minipage}[t]{0.44\linewidth}
    \centering
    \includegraphics[width=\linewidth]{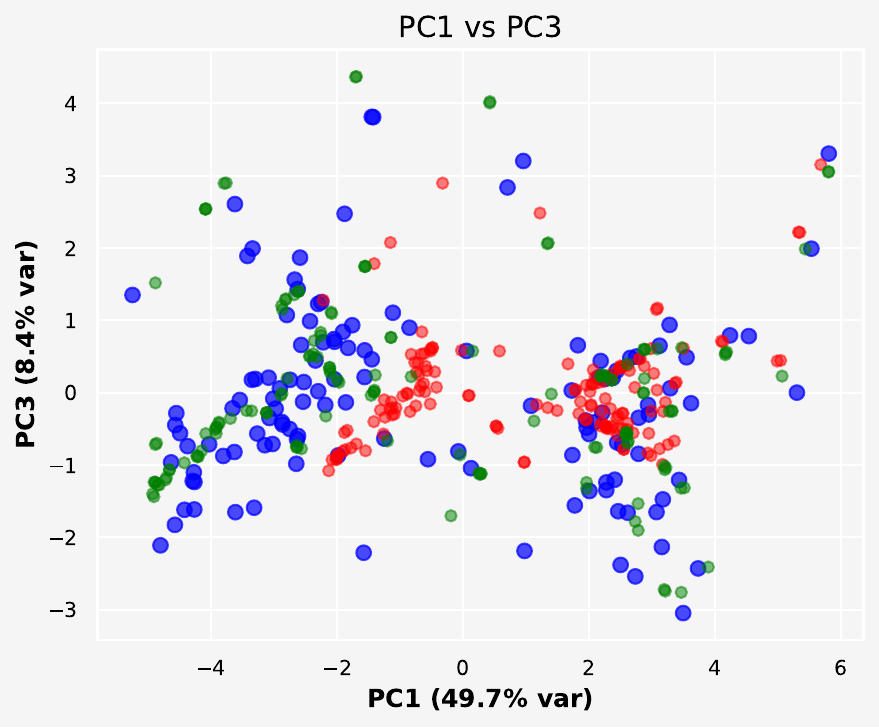}
    \par\smallskip
    {\scriptsize\textbf{(b)} PC1 vs PC3}
    \label{fig:pc1_pc3}
  \end{minipage}
  \begin{minipage}[t]{0.44\linewidth}
    \centering
    \includegraphics[width=\linewidth]{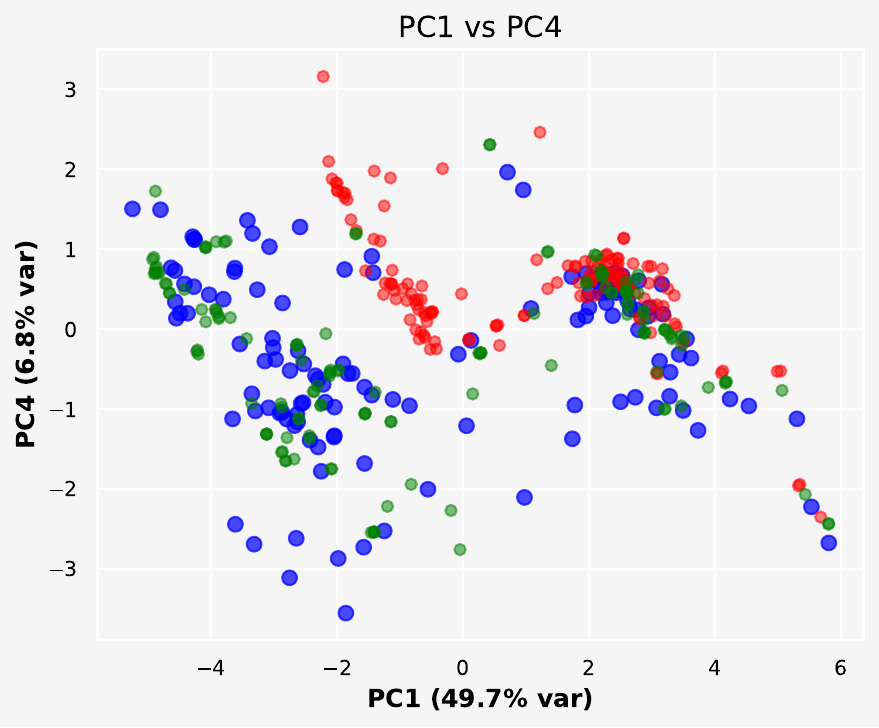}
    \par\smallskip
    {\scriptsize\textbf{(c)} PC1 vs PC4}
    \label{fig:pc1_pc4}
  \end{minipage}
  \begin{minipage}[t]{0.44\linewidth}
    \centering
    \includegraphics[width=\linewidth]{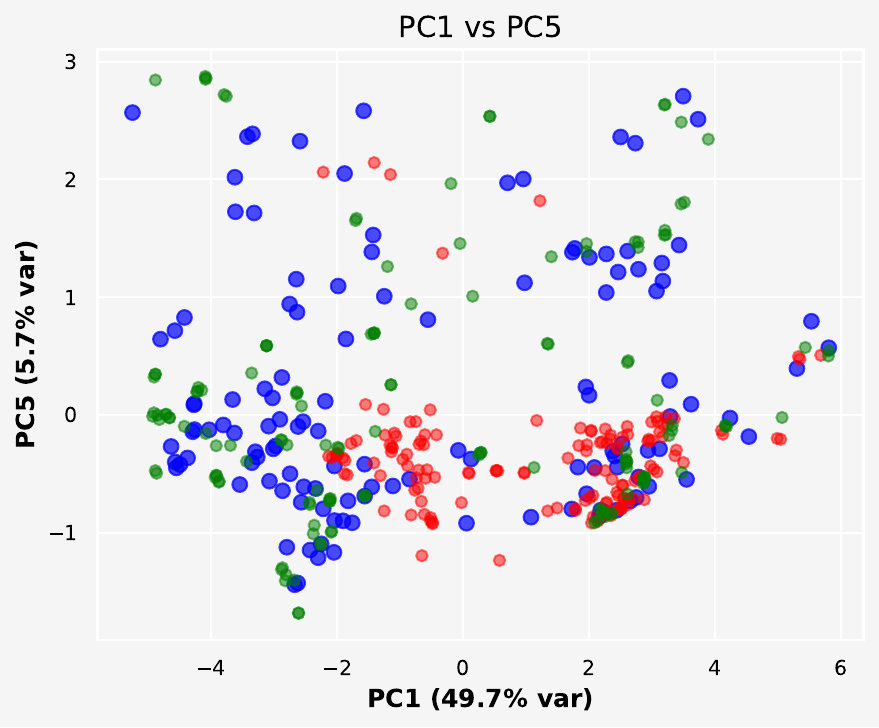}
    \par\smallskip
    {\scriptsize\textbf{(d)} PC1 vs PC5}
    \label{fig:pc1_pc5}
  \end{minipage}
  \caption{PCA projections of real, SMOGN-generated, and SMOGAN-refined samples (PC1 vs. PC2--PC5) for the Boston dataset. Blue, red, and green points correspond to real, SMOGN-generated, and SMOGAN-refined samples, respectively.}
  \label{fig:all_pca}
\end{figure}

\begin{figure}[!t]
  \centering
  \begin{minipage}[!t]{0.44\linewidth}
    \centering
    \includegraphics[width=\linewidth]{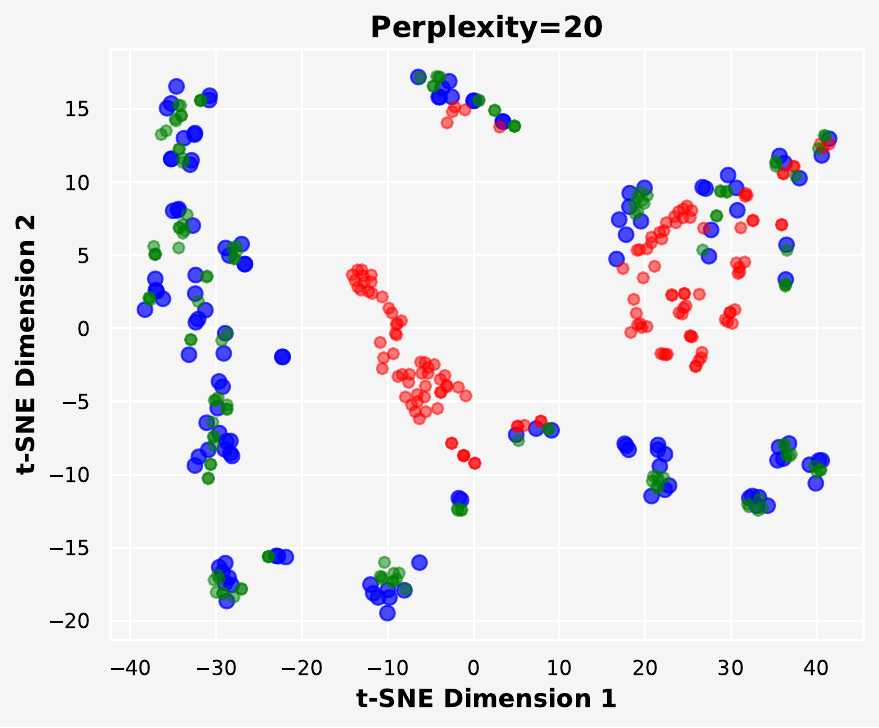}
    \par\smallskip
    {\scriptsize\textbf{(a)} Perplexity = 20}
    \label{fig:tsne_perp20}
  \end{minipage}
  \begin{minipage}[!t]{0.44\linewidth}
    \centering
    \includegraphics[width=\linewidth]{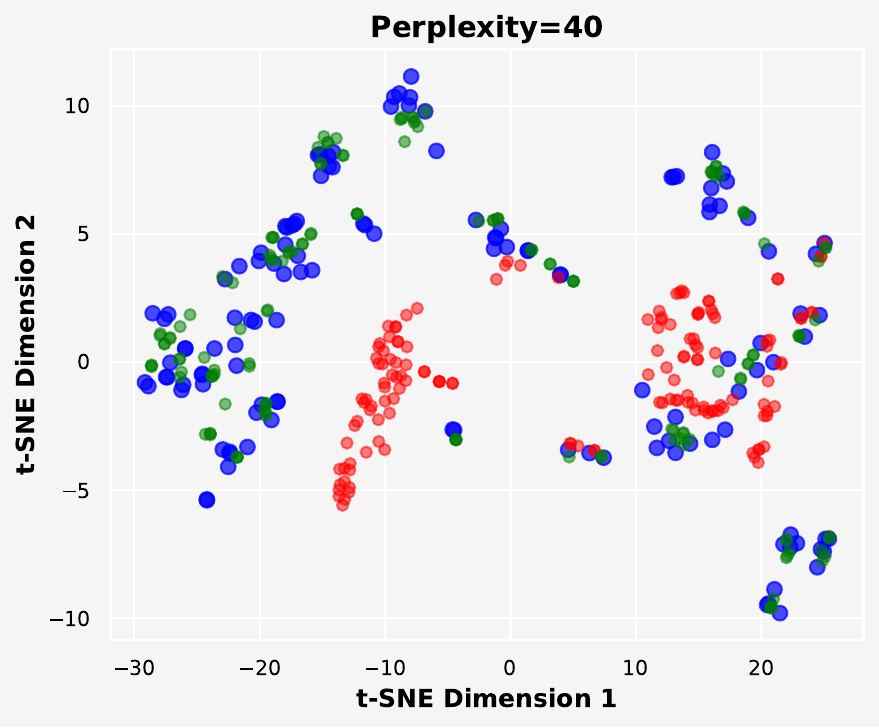}
    \par\smallskip
    {\scriptsize\textbf{(b)} Perplexity = 40}
    \label{fig:tsne_perp40}
  \end{minipage}
  \begin{minipage}[!t]{0.44\linewidth}
    \centering
    \includegraphics[width=\linewidth]{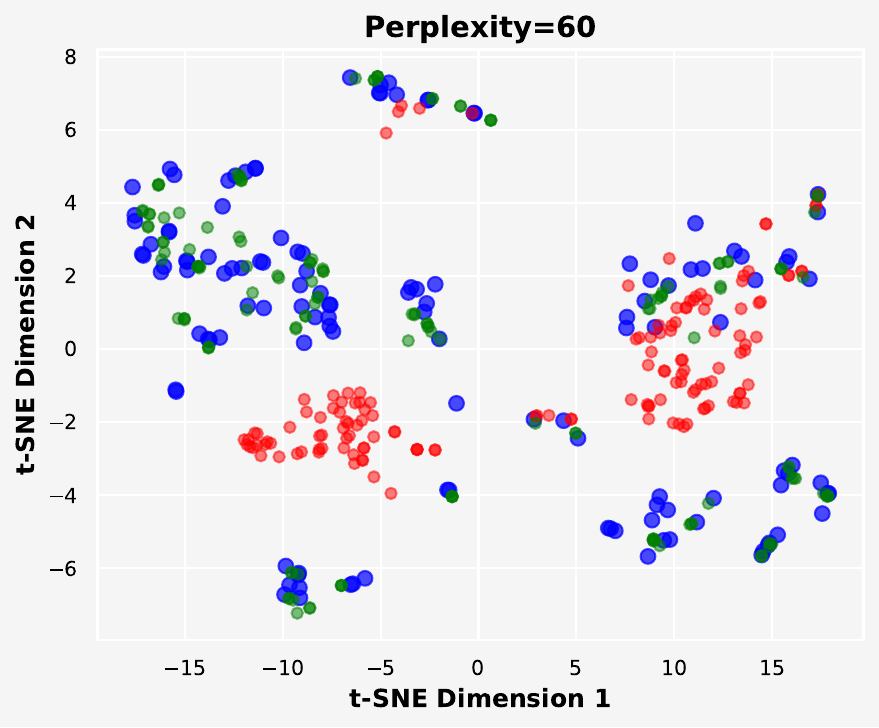}
    \par\smallskip
    {\scriptsize\textbf{(c)} Perplexity = 60}
    \label{fig:tsne_perp60}
  \end{minipage}
  \begin{minipage}[!t]{0.44\linewidth}
    \centering
    \includegraphics[width=\linewidth]{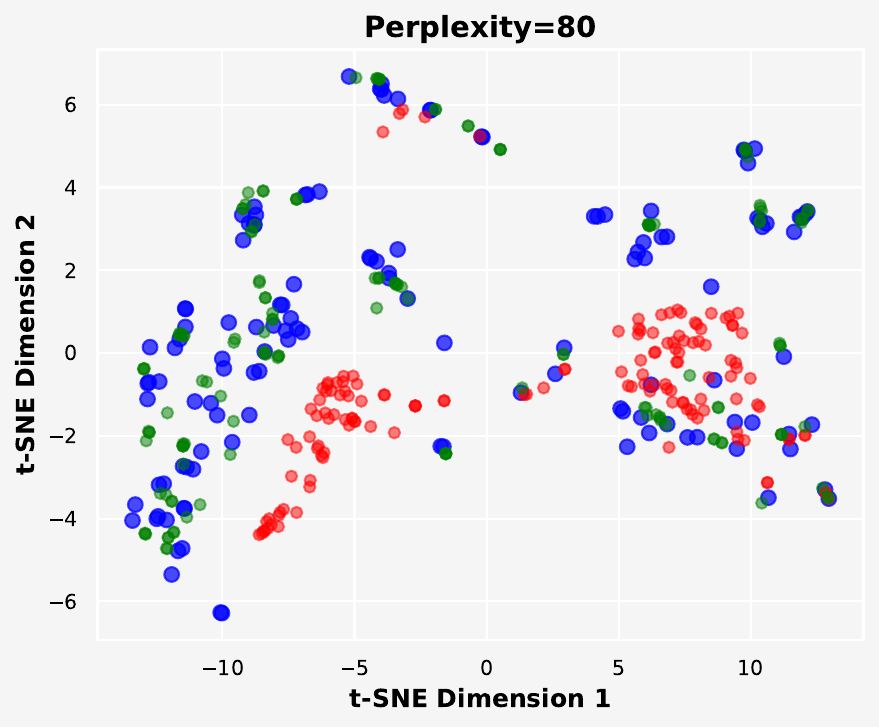}
    \par\smallskip
    {\scriptsize\textbf{(d)} Perplexity = 80}
    \label{fig:tsne_perp80}
  \end{minipage}
  \caption{t-SNE projections of real, SMOGN-generated, and SMOGAN-refined samples (perplexities 20, 40, 60, and 80) for the Boston dataset. Blue, red, and green points correspond to real, SMOGN-generated, and SMOGAN-refined samples, respectively.}
  \label{fig:all_tsne}
\end{figure}

\subsection{Principal component analysis}
Figure~\ref{fig:all_pca} presents PCA scatter plots of the original minority samples alongside SMOGN‑generated and SMOGAN‑refined samples for the Boston dataset. Each plot reports the explained variance of its principal components: PC1 alone accounts for 49.7\% of the variance, and the PC1 versus PC2 plot captures the greatest cumulative variance. The SMOGN initial synthetic samples (red) form a dense cluster and do not follow the natural spread of the original samples (blue). In contrast, the SMOGAN‑refined samples (green) are more widely distributed and more accurately reflect the true feature‑target relationship of the minority data.

\subsection{t‑Distributed stochastic neighbour embedding (t‑SNE)}

Figure~\ref{fig:all_tsne} presents t‑SNE scatter plots of the original minority samples alongside SMOGN‑generated and SMOGAN‑refined samples, using perplexities of 20, 40 and 60. Unlike PCA, which is a linear projection that maximizes global variance, t‑SNE is a nonlinear embedding that converts high‑dimensional similarities into probabilities and optimizes a 2D map by minimizing their Kullback-Leibler divergence \citep{vandermaaten2008visualizing}. Perplexity, roughly the effective number of nearest neighbours, balances local versus broader structure, and the axes “t‑SNE Dimension 1” and “t‑SNE Dimension 2” are simply the two coordinates of this embedding. As a result, t‑SNE highlights cluster separation and local density differences that PCA may obscure.

At perplexity 20, the original samples (blue) form several tight clusters, while the initial synthetic samples (red) collapse into a single dense blob. As perplexity increases to 40 and 60, the red points spread into multiple groups but still fail to align with the blue clusters. Even at perplexity 80, SMOGN samples remain poorly matched to the true manifold. In contrast, the SMOGAN‑refined samples (green) intermix with the blue points across all perplexities, showing that adversarial refinement recovers the local structure that t‑SNE highlights.

\subsection{Feature Correlation Analysis}

\begin{figure}[!t]
  \centering
  \begin{minipage}[t]{0.48\linewidth}
    \centering
    \includegraphics[width=\linewidth]{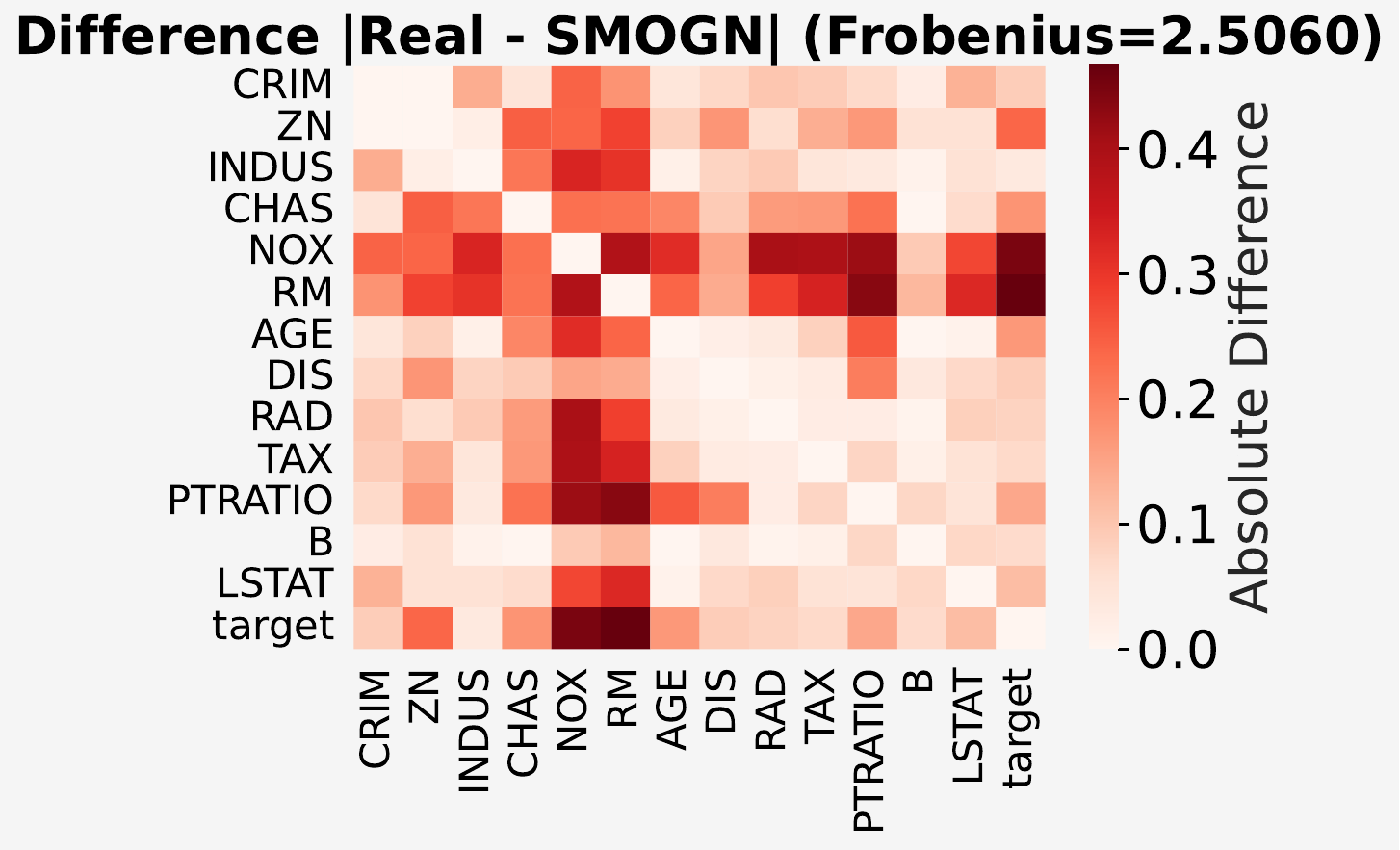}
    \par\smallskip
    {\scriptsize\textbf{(a)} Difference \(|\text{Real} - \text{SMOGN}|\)}
    \label{fig:diff_synth}
  \end{minipage}\hfill
  \begin{minipage}[t]{0.48\linewidth}
    \centering
    \includegraphics[width=\linewidth]{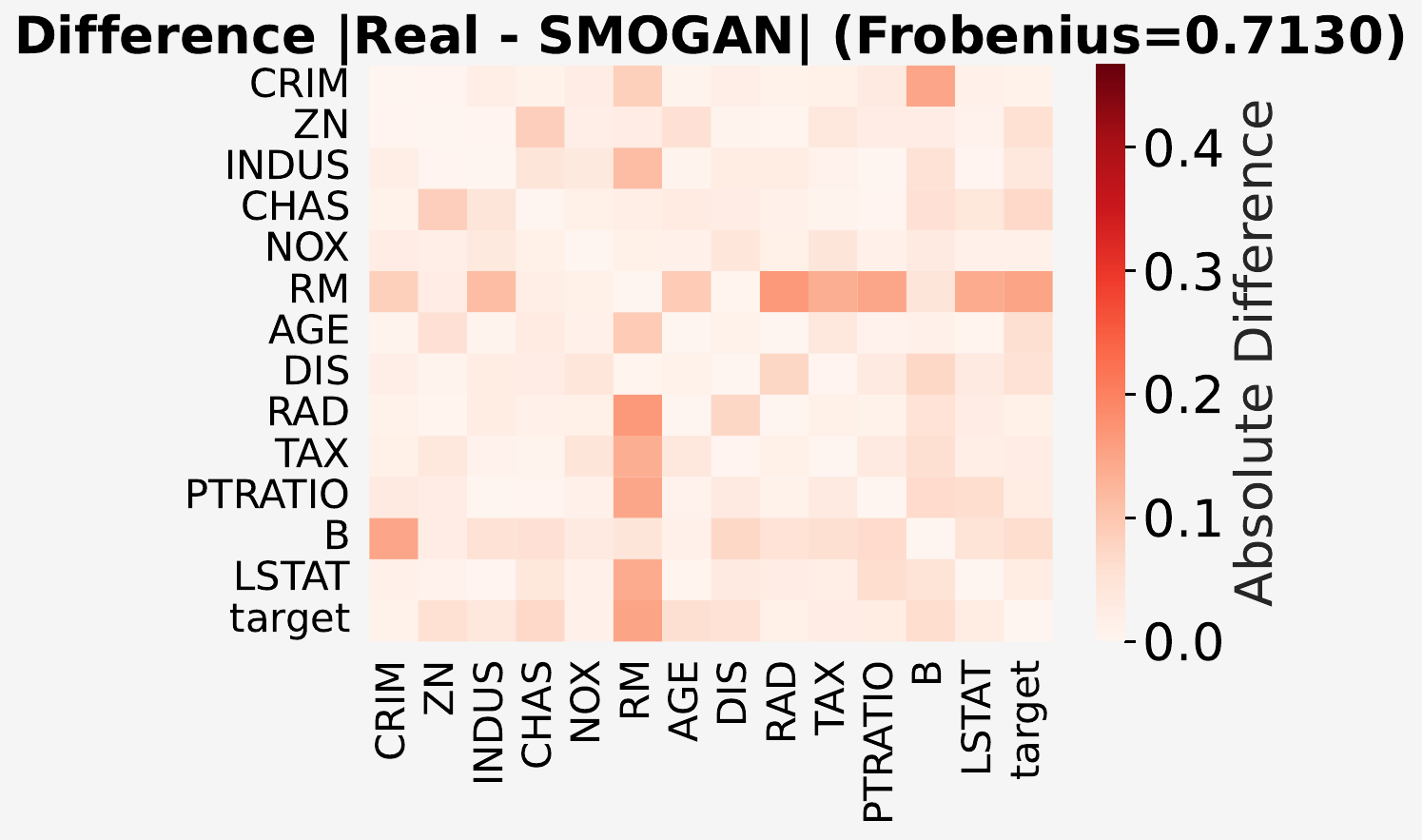}
    \par\smallskip
    {\scriptsize\textbf{(b)} Difference \(|\text{Real} - \text{SMOGAN}|\)}
    \label{fig:diff_refined}
  \end{minipage}
  
  \caption{Absolute difference between correlation matrices of real data and synthetic data generated by SMOGN (left) and SMOGAN (right).}
  \label{fig:diff_matrices}
\end{figure}

We compute Pearson correlation matrices for the real minority samples, SMOGN-generated samples, and SMOGAN-refined samples, considering both features and the target. Each matrix captures pairwise linear relationships among 13 features of the Boston dataset (including the target). To quantify how closely each synthetic method reproduces the true inter-feature structure, we take the element-wise absolute difference between the real and synthetic correlation matrices and compute their Frobenius norms:
\[
  \|\,C_{\text{real}} - C_{\text{method}}\,\|_F.
\]
Figure~\ref{fig:diff_matrices} visualizes the absolute difference heatmaps between the real minority correlation matrix and those of SMOGN (a) and SMOGAN (b). A lower Frobenius norm indicates a closer match to the real data’s correlation structure. In our experiments, the norm for the \(|\text{Real} - \text{SMOGN}|\) matrix is 2.5060 versus 0.7130 for \(|\text{Real} - \text{SMOGAN}|\), representing a 72\% reduction in overall discrepancy. This substantial drop shows that SMOGAN’s adversarial refinement preserves inter‑feature relationships far better than the initial SMOGN oversampler.

\subsection{Statistical Property Analysis}

We compute four univariate statistics, mean, standard deviation, skewness, and kurtosis, for each feature (including the target) on the real minority samples and on the synthetic samples generated by SMOGN and refined by SMOGAN. For each statistic, we calculate the absolute difference between the real and synthetic values and then average those differences across all features. A lower absolute difference indicates that the synthetic data more faithfully reproduces the marginal distribution of the real data.

\begin{figure}[!t]
  \centering
  \includegraphics[width=0.8\linewidth, height=6cm, keepaspectratio=false]{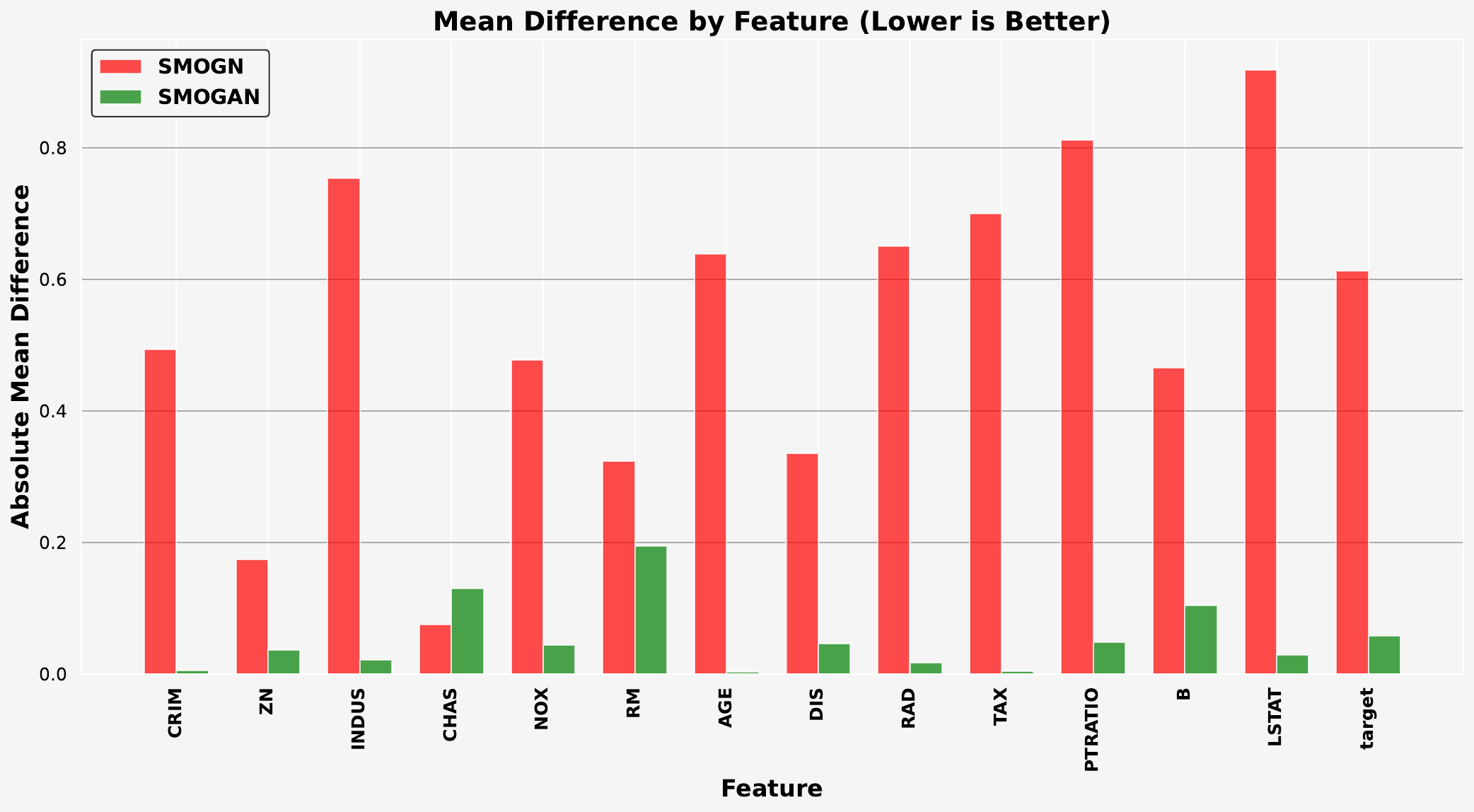}
  \caption{Absolute mean differences for each feature (lower is better). Red bars correspond to SMOGN, and green bars (SMOGAN) show the refined method's improvement.}
  \label{fig:stat_props_bar}
\end{figure}

Figure~\ref{fig:stat_props_bar} shows the absolute mean differences by feature for both SMOGN and SMOGAN, with shorter bars indicating closer alignment of the synthetic sample means with the real data, and Table~\ref{table:stat_props_summary} provides the numeric summary of average differences and percentage improvements.

\begin{table}[!t]
  \caption{Average absolute differences and improvement (\%) for univariate statistics.}
  \label{table:stat_props_summary}
  \centering
  \small
  \setlength{\tabcolsep}{4pt}
  \begin{tabular}{lrrr}
    \toprule
    Statistic           & SMOGN diff & SMOGAN diff & Imp. (\%) \\
    \midrule
    Mean                & 0.5309     & \textbf{0.0530} & 90.02      \\
    Standard deviation  & 0.4495     & \textbf{0.0866} & 80.74      \\
    Skewness            & 0.7525     & \textbf{0.1246} & 83.44      \\
    Kurtosis            & 2.1472     & \textbf{0.5270} & 75.46      \\
    \bottomrule
  \end{tabular}
\end{table}

\begin{figure*}[!t]
  \centering
  \includegraphics[width=\columnwidth]{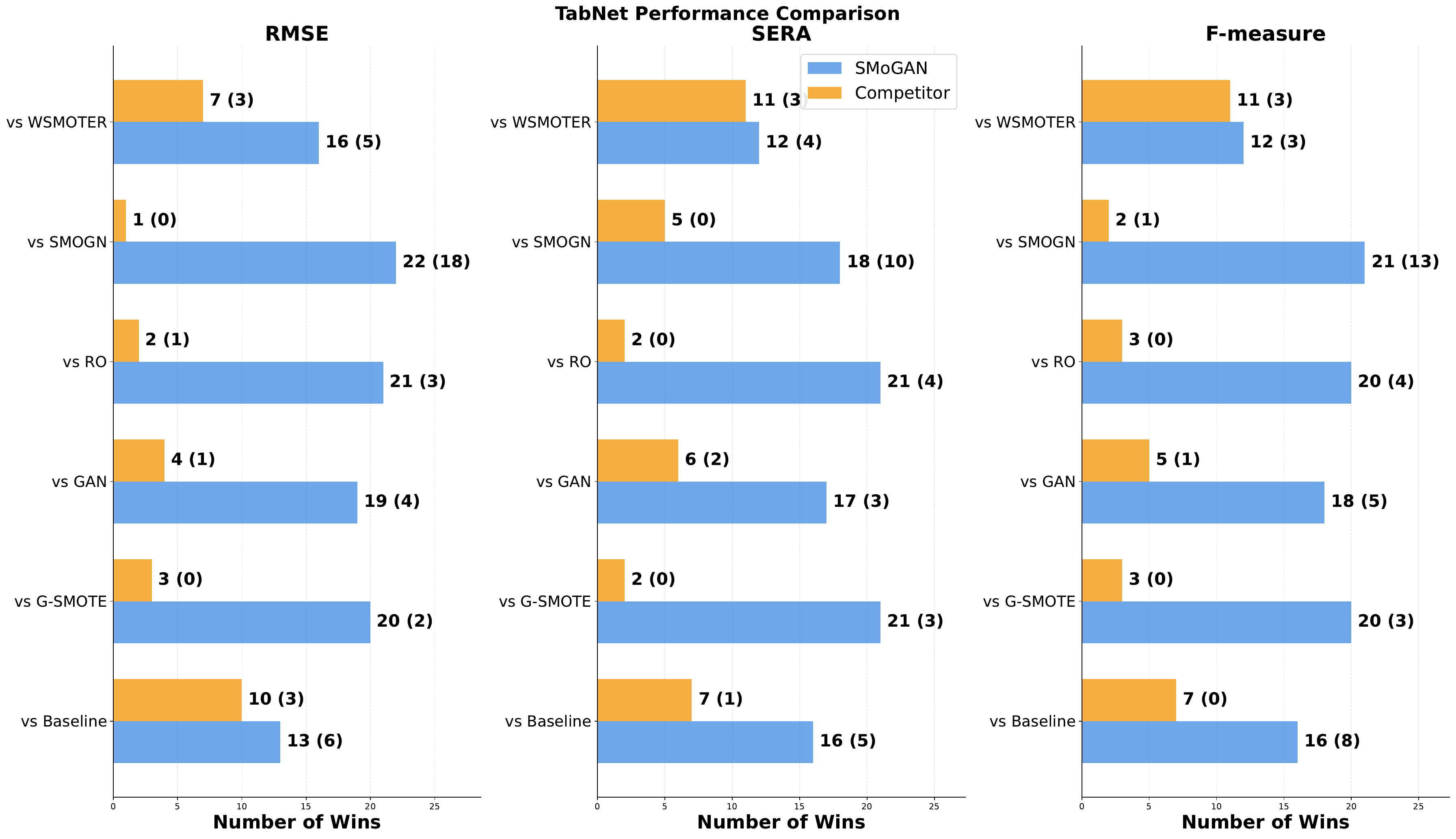}
  \caption{SMoGAN performance comparison across methods and metrics for TabNet model. Blue bars represent SMoGAN wins, orange bars represent competitor wins. Numbers show total wins with significant wins in parentheses. Higher values indicate better performance.}
  \label{fig:tabnet}
\end{figure*}

\begin{figure*}[!t]
  \centering
  \includegraphics[width=\columnwidth]{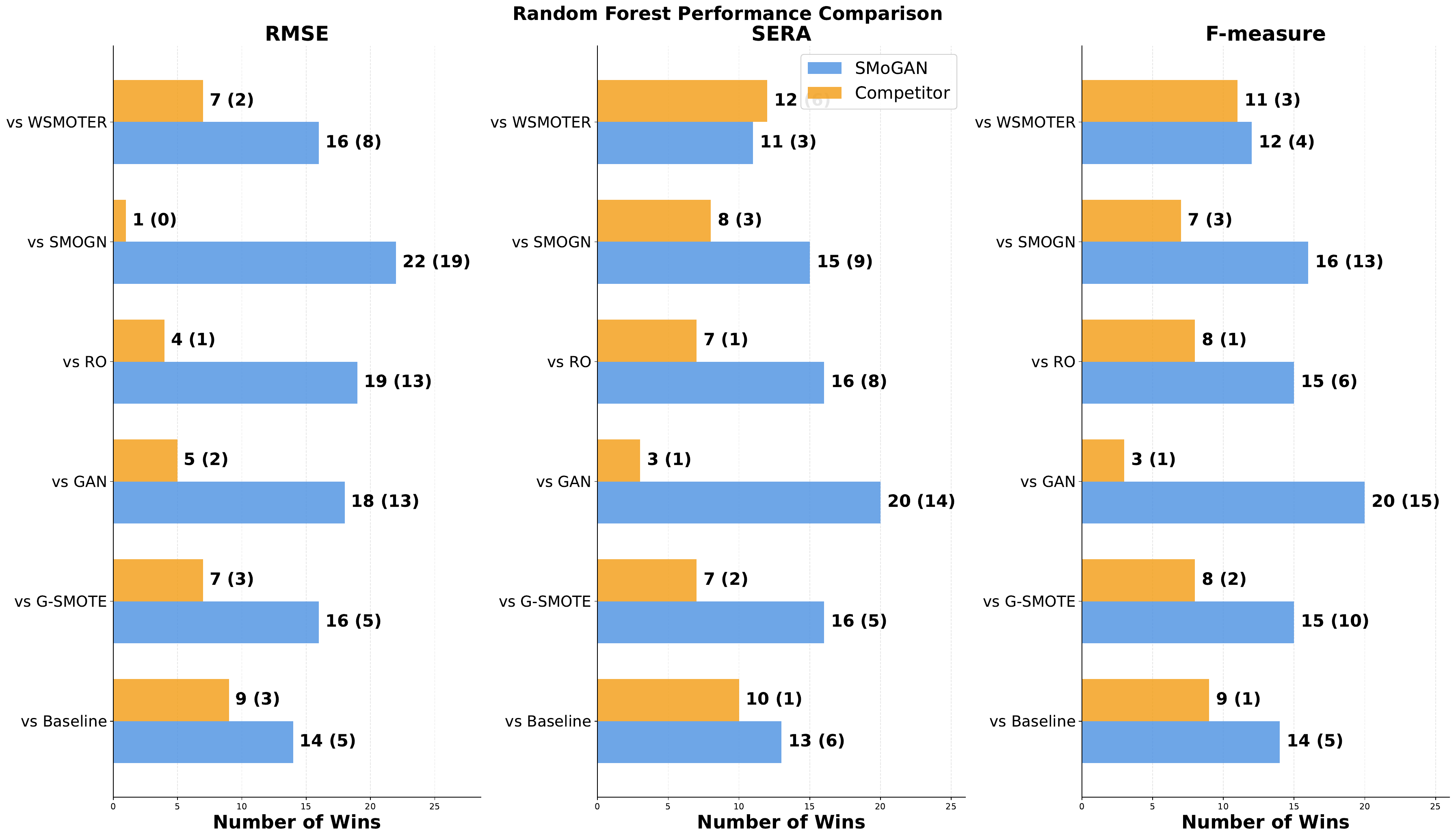}
  \caption{SMoGAN performance comparison across methods and metrics for random forest model. Blue bars represent SMoGAN wins, orange bars represent competitor wins. Numbers show total wins with significant wins in parentheses. Higher values indicate better performance.}
  \label{fig:rf}
\end{figure*}

\section{Results}

We present the results of comparing SMOGAN with the state-of-the-art oversamplers, SMOGN, WSMOTER, RO and G-SMOTE, as well as baseline (no oversampling) and GAN-only approach across various performance dimensions. Note that the SMOGN dataset comprises the original data plus synthetic samples generated by SMOGN, while the SMOGAN dataset comprises the original data plus refined synthetic samples after DistGAN filtration. We conduct aggregated pairwise comparisons at relevance thresholds \(t_R = 0.8\) for each evaluation metric (RMSE, SERA, F-measure).

For each dataset and each pairwise comparison, a winner is determined based on outcomes across 25 random splits, so each dataset contributes one win per threshold and two wins in total across both thresholds. Specifically, for any two methods under comparison (for example, SMOGAN vs.\ SMOGN), the method that prevails in the majority of splits is declared the winner. Next, we perform the Wilcoxon signed-rank test for each pairwise comparison. We compute the differences between the 25-split metric values of the two methods to form a difference vector, and then the test assesses whether these differences are statistically significant at an alpha level of 0.05, corresponding to a 5\% risk of incorrectly concluding that a difference exists when there is none \citep{wilcoxon1945}.

In Figures~\ref{fig:tabnet} and \ref{fig:rf}, the numbers outside the parentheses indicate the total number of wins for each method, while the numbers inside the parentheses represent statistically significant wins. SMOGAN outperforms all competitors across all three metrics (RMSE, SERA, and F-measure) for both TabNet and Random Forest (except against WSMOTER for SERA in the Random Forest model). It not only records a higher total number of wins in every comparison but also contributes the majority of statistically significant wins, demonstrating that adversarial refinement produces synthetic samples of substantially higher fidelity and SMOGAN's refined data yield consistently better alignment with the true distribution and thus stronger predictive accuracy.

Two key comparisons highlight SMOGAN's effectiveness. First, SMOGAN significantly outperforms the GAN-only method, demonstrating that it is not simply the GAN process that generates suitable data. Providing the GAN with SMOGN-generated starting points enables more consistent synthetic data generation compared to purely adversarial approaches. Second, SMOGAN's superiority over SMOGN alone shows that the adversarial loss function successfully moves the synthetic data toward more realistic distributions. This confirms the evidence presented in Section 5, where SMOGAN data demonstrates higher fidelity reproduction of natural minority patterns compared to traditional oversampling methods.

\section{Limitations and Discussion}
SMOGAN’s adversarial refinement adds complexity and computational overhead and requires fine-tuning of GAN-specific hyperparameters for optimal results. Future work will focus on automated fine-tuning strategies for relevance thresholds, more efficient adversarial training schedules, and extensions to high-dimensional or multimodal datasets. Overall, SMOGAN offers a general, data-driven approach that improves the fidelity and utility of synthetic samples for imbalanced regression tasks.

\section{Conclusion}

Our experiments show that even when user-selected thresholds are used to define minority samples, SMOGAN enhances generalization and predictive accuracy, particularly for extreme values, by generating high-quality synthetic data. In practice, SMOGAN can be integrated into any data-level oversampling pipeline to improve model performance. SMOGAN’s filtering framework addresses low-quality synthetic data by integrating our DistGAN adversarial filtering layer that aligns generated samples with the true joint feature-target distribution. Qualitative analyses (PCA, t-SNE, correlation alignment, statistical moments) and quantitative benchmarks (RMSE, SERA, F-measure) consistently demonstrate that SMOGAN produces more realistic data and achieves superior predictive performance compared to the base oversampler and the baseline model.

\bibliographystyle{plainnat}
\bibliography{references}

\begin{thebibliography}{48}
\providecommand{\natexlab}[1]{#1}
\providecommand{\url}[1]{\texttt{#1}}
\expandafter\ifx\csname urlstyle\endcsname\relax
  \providecommand{\doi}[1]{doi: #1}\else
  \providecommand{\doi}{doi: \begingroup \urlstyle{rm}\Url}\fi

\bibitem[Alahyari and Domaratzki(2025)]{alahyari2025ldao}
Shayan Alahyari and Mike Domaratzki.
\newblock Local distribution-based adaptive oversampling for imbalanced regression, April 2025.

\bibitem[Alahyari et~al.(2025)Alahyari, Ghobadlou, and Domaratzki]{alahyari2025regression}
Shayan Alahyari, Shiva~Mehdipour Ghobadlou, and Mike Domaratzki.
\newblock Regression augmentation with data-driven segmentation, August 2025.

\bibitem[Alcalá-Fdez et~al.(2011)Alcalá-Fdez, Fernandez, Luengo, Derrac, García, Sánchez, and Herrera]{alcala2011}
J.~Alcalá-Fdez, A.~Fernandez, J.~Luengo, J.~Derrac, S.~García, L.~Sánchez, and F.~Herrera.
\newblock Keel data-mining software tool: Data set repository, integration of algorithms and experimental analysis framework.
\newblock \emph{Journal of Multiple-Valued Logic and Soft Computing}, 17\penalty0 (2-3):\penalty0 255--287, 2011.

\bibitem[Ar{\i}k and Pfister(2021)]{arik2021tabnet}
Sercan~{\"O}. Ar{\i}k and Tomas Pfister.
\newblock Tabnet: Attentive interpretable tabular learning.
\newblock In \emph{Proceedings of the AAAI Conference on Artificial Intelligence}, volume~35, pages 6679--6687, 2021.
\newblock \doi{10.1609/aaai.v35i8.16826}.

\bibitem[Arjovsky et~al.(2017)Arjovsky, Chintala, and Bottou]{arjovsky2017wasserstein}
Martin Arjovsky, Soumith Chintala, and Léon Bottou.
\newblock {Wasserstein} {GAN}.
\newblock \emph{arXiv preprint arXiv:1701.07875v3 [stat.ML]}, 2017.

\bibitem[Avelino et~al.(2024)Avelino, Cavalcanti, and Cruz]{avelino2024}
J.~G. Avelino, G.~D.~C. Cavalcanti, and R.~M.~O. Cruz.
\newblock Resampling strategies for imbalanced regression: a survey and empirical analysis.
\newblock \emph{Artificial Intelligence Review}, 57:\penalty0 Article 82, 2024.

\bibitem[Branco et~al.(2016)Branco, Torgo, and Ribeiro]{branco2016}
P.~Branco, L.~Torgo, and R.~P. Ribeiro.
\newblock A survey of predictive modeling under imbalanced distributions.
\newblock \emph{ACM Computing Surveys}, 49\penalty0 (2):\penalty0 Article 31, 2016.

\bibitem[Branco et~al.(2017)Branco, Torgo, and Ribeiro]{branco2017}
P.~Branco, L.~Torgo, and R.~P. Ribeiro.
\newblock Smogn: A pre-processing approach for imbalanced regression.
\newblock In \emph{Proceedings of Machine Learning Research: LIDTA}, volume~74, pages 36--50, 2017.

\bibitem[Branco et~al.(2019)Branco, Torgo, and Ribeiro]{branco2019}
P.~Branco, L.~Torgo, and R.~P. Ribeiro.
\newblock Pre-processing approaches for imbalanced distributions in regression.
\newblock \emph{Neurocomputing}, 343:\penalty0 76--99, 2019.

\bibitem[Breiman(2001)]{breiman2001random}
Leo Breiman.
\newblock Random forests.
\newblock \emph{Machine Learning}, 45\penalty0 (1):\penalty0 5--32, 2001.
\newblock \doi{10.1023/A:1010933404324}.

\bibitem[Buda et~al.(2018)Buda, Maki, and Mazurowski]{buda2018}
M.~Buda, A.~Maki, and M.~A. Mazurowski.
\newblock A systematic study of the class imbalance problem in convolutional neural networks.
\newblock \emph{Neural Networks}, 106:\penalty0 249--259, 2018.

\bibitem[Camacho and Bacao(2024)]{camacho2024}
L.~Camacho and F.~Bacao.
\newblock {WSMOTER}: A novel approach for imbalanced regression.
\newblock \emph{Applied Intelligence}, 54:\penalty0 8789--8799, 2024.

\bibitem[Camacho et~al.(2022)Camacho, Douzas, and Bacao]{camacho2022}
L.~Camacho, G.~Douzas, and F.~Bacao.
\newblock Geometric {SMOTE} for regression.
\newblock \emph{Expert Systems with Applications}, 193:\penalty0 116387, 2022.

\bibitem[Chawla et~al.(2002)Chawla, Bowyer, Hall, and Kegelmeyer]{chawla2002}
N.~V. Chawla, K.~W. Bowyer, L.~O. Hall, and W.~P. Kegelmeyer.
\newblock {SMOTE}: Synthetic minority over-sampling technique.
\newblock \emph{Journal of Artificial Intelligence Research}, 16:\penalty0 321--357, 2002.

\bibitem[Chawla et~al.(2004)Chawla, Japkowicz, and Kolcz]{chawla2004}
N.~V. Chawla, N.~Japkowicz, and A.~Kolcz.
\newblock Editorial: Special issue on learning from imbalanced data sets.
\newblock \emph{ACM SIGKDD Explorations Newsletter}, 6\penalty0 (1):\penalty0 1--6, 2004.

\bibitem[Domingos(1999)]{domingos1999}
P.~Domingos.
\newblock Metacost: A general method for making classifiers cost-sensitive.
\newblock In \emph{Proceedings of the 5th ACM SIGKDD International Conference on Knowledge Discovery and Data Mining (KDD)}, pages 155--164, 1999.

\bibitem[Elkan(2001)]{elkan2001}
C.~Elkan.
\newblock The foundations of cost-sensitive learning.
\newblock In \emph{Proceedings of the 17th International Joint Conference on Artificial Intelligence (IJCAI)}, pages 973--978, 2001.

\bibitem[Engelmann and Lessmann(2020)]{engelmann2020conditional}
Justin Engelmann and Stefan Lessmann.
\newblock Conditional wasserstein {GAN}‐based oversampling of tabular data for imbalanced learning.
\newblock Preprint arXiv:2008.09202v1, 2020.

\bibitem[Goodfellow et~al.(2014)Goodfellow, Pouget‐Abadie, Mirza, Xu, Warde‐Farley, Ozair, Courville, and Bengio]{goodfellow2014generative}
Ian~J. Goodfellow, Jean Pouget‐Abadie, Mehdi Mirza, Bing Xu, David Warde‐Farley, Sherjil Ozair, Aaron Courville, and Yoshua Bengio.
\newblock Generative adversarial nets.
\newblock In \emph{Advances in Neural Information Processing Systems}, volume~27, pages 2672--2680, 2014.

\bibitem[Gulrajani et~al.(2017)Gulrajani, Ahmed, Arjovsky, Dumoulin, and Courville]{gulrajani2017improved}
Ishaan Gulrajani, Faruk Ahmed, Martin Arjovsky, Vincent Dumoulin, and Aaron Courville.
\newblock Improved training of {Wasserstein} {GANs}.
\newblock In \emph{Advances in Neural Information Processing Systems 30 (NeurIPS 2017)}, pages 5767--5777, 2017.

\bibitem[Guo et~al.(2017)Guo, Li, Shang, Gu, Huang, and Gong]{haixiang2017}
H.~Guo, Y.~Li, J.~Shang, M.~Gu, Y.~Huang, and B.~Gong.
\newblock Learning from class-imbalanced data: Review of methods and applications.
\newblock \emph{Expert Systems with Applications}, 73:\penalty0 220--239, 2017.

\bibitem[He and Garcia(2009)]{he2009}
H.~He and E.~A. Garcia.
\newblock Learning from imbalanced data.
\newblock \emph{IEEE Transactions on Knowledge and Data Engineering}, 21\penalty0 (9):\penalty0 1263--1284, 2009.

\bibitem[Jiang et~al.(2019)Jiang, Hong, Zhou, He, and Cheng]{jiang2019gan}
Wenqian Jiang, Yang Hong, Beitong Zhou, Xin He, and Cheng Cheng.
\newblock A {GAN}‐based anomaly detection approach for imbalanced industrial time series.
\newblock \emph{IEEE Access}, 7:\penalty0 143608--143619, 2019.

\bibitem[Johnson and Khoshgoftaar(2019)]{johnson2019}
J.~M. Johnson and T.~M. Khoshgoftaar.
\newblock Survey on deep learning with class imbalance.
\newblock \emph{Journal of Big Data}, 6\penalty0 (1):\penalty0 1--54, 2019.

\bibitem[Kamangir et~al.(2024)Kamangir, Sams, Dokoozlian, Sanchez, and Earles]{kamangir2024large}
Hossein Kamangir, S.~Sams, B.\, Nuno Dokoozlian, Luis Sanchez, and M.~Earles, J.\.
\newblock Large-scale spatio-temporal yield estimation via deep learning using satellite and management data fusion in vineyards.
\newblock \emph{Computers and Electronics in Agriculture}, 216:\penalty0 108439, 2024.

\bibitem[Krawczyk(2016)]{krawczyk2016}
B.~Krawczyk.
\newblock Learning from imbalanced data: open challenges and future directions.
\newblock \emph{Progress in Artificial Intelligence}, 5\penalty0 (4):\penalty0 221--232, 2016.

\bibitem[Kunz(2020)]{kunz2020}
N.~Kunz.
\newblock Smogn: Synthetic minority over-sampling technique for regression with {Gaussian} noise.
\newblock PyPI, version v0.1.2, 2020.

\bibitem[Lee and Park(2021)]{lee2019ganids}
JooHwa Lee and KeeHyun Park.
\newblock {GAN}-based imbalanced data intrusion detection system.
\newblock \emph{Personal and Ubiquitous Computing}, 25\penalty0 (1):\penalty0 121--128, 2021.

\bibitem[Liu et~al.(2009)Liu, Wu, and Zhou]{liu2009}
X.-Y. Liu, J.~Wu, and Z.-H. Zhou.
\newblock Exploratory undersampling for class-imbalance learning.
\newblock \emph{IEEE Transactions on Systems, Man, and Cybernetics, Part B (Cybernetics)}, 39\penalty0 (2):\penalty0 539--550, 2009.

\bibitem[Ma et~al.(2024)Ma, Huang, Nan, Moniz, Zhang, Wiest, and Chawla]{ma2024revisiting}
Yihong Ma, Xiaobao Huang, Bozhao Nan, Nuno Moniz, Xiangliang Zhang, Olaf Wiest, and Nitesh~V. Chawla.
\newblock Are we making much progress? revisiting chemical reaction yield prediction from an imbalanced regression perspective.
\newblock In \emph{Companion Proceedings of the ACM Web Conference 2024 (WWW '24 Companion)}, pages 791--794. ACM, 2024.

\bibitem[Mariani et~al.(2018)Mariani, Scheidegger, Istrate, Bekas, and Malossi]{mariani2018bagan}
Giovanni Mariani, Florian Scheidegger, Roxana Istrate, Costas Bekas, and Cristiano Malossi.
\newblock {BAGAN}: Data augmentation with balancing {GAN}.
\newblock Preprint arXiv:1803.09655v2, 2018.

\bibitem[Moniz et~al.(2018)Moniz, Torgo, and Soares]{moniz2018}
N.~Moniz, L.~Torgo, and C.~Soares.
\newblock {SMOTEBoost} for regression: Improving the prediction of extreme values.
\newblock In \emph{Proceedings of the 5th International Conference on Data Science and Advanced Analytics (DSAA)}, pages 127--136, 2018.

\bibitem[Radovanovi{\'c} et~al.(2025)Radovanovi{\'c}, Rafiei, Grunwell, and Kamaleswaran]{radovanovic2025tackling}
Valentina Radovanovi{\'c}, Alireza Rafiei, Jocelyn Grunwell, and Rishikesan Kamaleswaran.
\newblock Tackling small imbalanced regression datasets by stability selection and smogn: A case study of ventilation-free days prediction in a pediatric intensive care unit.
\newblock \emph{International Journal of Medical Informatics}, 196:\penalty0 105809, 2025.

\bibitem[Ren et~al.(2022)Ren, Luo, and Urtasun]{ren2022}
M.~Ren, W.~Luo, and R.~Urtasun.
\newblock Balanced mse for imbalanced visual regression.
\newblock In \emph{Proceedings of the IEEE/CVF Conference on Computer Vision and Pattern Recognition (CVPR)}, pages 418--427, 2022.

\bibitem[Ribeiro and Moniz(2020)]{ribeiro2020}
R.~P. Ribeiro and N.~Moniz.
\newblock Imbalanced regression and extreme value prediction.
\newblock \emph{Machine Learning}, 109\penalty0 (9-10):\penalty0 1803--1835, 2020.

\bibitem[Ribeiro(2011)]{ribeiro2011a}
R.~P.~A. Ribeiro.
\newblock \emph{Utility-based regression}.
\newblock PhD thesis, Faculty of Sciences, University of Porto, Porto, 2011.

\bibitem[Sharma et~al.(2022)Sharma, Singh, and Chandra]{sharma2022smotified}
Anuraganand Sharma, Prabhat~Kumar Singh, and Rohitash Chandra.
\newblock {SMOTified}-{GAN} for class imbalanced pattern classification problems.
\newblock \emph{IEEE Access}, 10:\penalty0 30655--30665, 2022.

\bibitem[Steininger et~al.(2021)Steininger, Kobs, Davidson, Krause, and Hotho]{steininger2021}
M.~Steininger, K.~Kobs, P.~Davidson, A.~Krause, and A.~Hotho.
\newblock Density-based weighting for imbalanced regression.
\newblock \emph{Machine Learning}, 110\penalty0 (8):\penalty0 2187--2210, 2021.

\bibitem[Tanaka and Aranha(2019)]{tanaka2019dataaugmentation}
Fabio Henrique Kiyoiti dos~Santos Tanaka and Claus Aranha.
\newblock Data augmentation using {GANs}.
\newblock In \emph{Proceedings of Machine Learning Research}, volume XXX, pages 1--16, 2019.

\bibitem[Torgo et~al.(2013)Torgo, Ribeiro, da~Costa, and Pal]{torgo2013}
L.~Torgo, R.~P. Ribeiro, J.~P. da~Costa, and S.~Pal.
\newblock {SMOTE} for regression.
\newblock In \emph{Intelligent Data Engineering and Automated Learning (IDEAL 2013). Lecture Notes in Computer Science}, volume 8206, pages 378--387, 2013.

\bibitem[Torgo and Ribeiro(2007)]{torgo2007utility}
Luis Torgo and Rita Ribeiro.
\newblock Utility-based regression.
\newblock In \emph{Proceedings of the 11th European Conference on Principles and Practice of Knowledge Discovery in Databases (PKDD 2007)}, pages 597--604, 2007.

\bibitem[Torgo and Ribeiro(2009)]{torgo2009}
Luis Torgo and Rita Ribeiro.
\newblock Precision and recall for regression.
\newblock In \emph{Discovery Science (DS 2009)}, volume 5808 of \emph{Lecture Notes in Artificial Intelligence}, pages 332--346. Springer‑Verlag Berlin Heidelberg, 2009.

\bibitem[van~der Maaten and Hinton(2008)]{vandermaaten2008visualizing}
Laurens van~der Maaten and Geoffrey Hinton.
\newblock Visualizing data using t-sne.
\newblock \emph{Journal of Machine Learning Research}, 9:\penalty0 2579--2605, 2008.

\bibitem[Wilcoxon(1945)]{wilcoxon1945}
F.~Wilcoxon.
\newblock Individual comparisons by ranking methods.
\newblock \emph{Biometrics Bulletin}, 1\penalty0 (6):\penalty0 80--83, 1945.

\bibitem[Wu et~al.(2022)Wu, Kunz, and Branco]{wu2022}
W.~Wu, N.~Kunz, and P.~Branco.
\newblock Imbalancedlearningregression-a python package to tackle the imbalanced regression problem.
\newblock In \emph{Joint European Conference on Machine Learning and Knowledge Discovery in Databases}, pages 645--648, 2022.

\bibitem[Yang et~al.(2021)Yang, Xie, Yu, He, and Liu]{yang2021}
J.~Yang, L.~Xie, Q.~Yu, X.~He, and J.~Liu.
\newblock Delving into deep imbalanced regression.
\newblock In \emph{Proceedings of the 38th International Conference on Machine Learning (ICML)}, pages 8437--8447, 2021.

\bibitem[Zhen et~al.(2025)Zhen, Chen, Wang, Yang, Xu, and Huang]{zhen2025weighted}
Ling Zhen, Baihua Chen, Lin Wang, Lin Yang, Wei Xu, and Ru-Jin Huang.
\newblock Weighted support vector regression for high ozone concentration forecasting.
\newblock \emph{Atmospheric Environment}, 343:\penalty0 120952, 2025.

\bibitem[Zhou and Liu(2010)]{zhou2010}
Z.-H. Zhou and X.-Y. Liu.
\newblock On multi-class cost-sensitive learning.
\newblock \emph{Computational Intelligence}, 26\penalty0 (3):\penalty0 232--257, 2010.

\end{thebibliography}

\end{document}